\pdfoutput=1

\documentclass[11pt]{article}
\usepackage[table]{xcolor}

\usepackage[preprint]{acl}

\usepackage{times}
\usepackage{latexsym}
\usepackage{enumitem}
\usepackage[T1]{fontenc}

\usepackage[utf8]{inputenc}

\usepackage{microtype}
\usepackage{makecell}
\usepackage{bbding}
\usepackage{inconsolata}

\usepackage{graphicx}
\usepackage{subcaption}
\usepackage{caption}
\usepackage{multirow}
\usepackage{amsmath,amssymb}
\usepackage{booktabs}
\usepackage[normalem]{ulem}

\title{Could Thinking Multilingually Empower LLM Reasoning?}

\newcommand*{\affaddr}[1]{#1} 
\newcommand*{\affmark}[1][*]{\textsuperscript{#1}}
\newcommand*{\email}[1]{\texttt{#1}}

\author{%
Changjiang Gao\affmark[1]\text{,} Xu Huang\affmark[1]\text{,} Wenhao Zhu\affmark[1]\text{,} Shujian Huang\affmark[1]\thanks{Corresponding authors.}\text{,} Lei Li\affmark[3]\text{,} Fei Yuan\affmark[2]\footnotemark[\value{footnote}] \\
\affaddr{\affmark[1]National Key Laboratory for Novel Software Technology, Nanjing University}\\
\affaddr{\affmark[2]Shanghai Artificial Intelligence Laboratory}, \affaddr{\affmark[3]Carnegie Mellon University} \\
\email{\{gaocj,xuhuang,zhuwh\}@smail.nju.edu.cn}, \email{huangsj@nju.edu.cn} \\
\email{leili@cs.cmu.edu}, \email{yuanfei@pjlab.org.cn}\\
}

\begin{document}
\maketitle

\begin{abstract}


Previous work indicates that large language models exhibit a significant ``English bias'', i.e. they often perform better when tasks are presented in English. Interestingly, we have observed that using certain other languages in reasoning tasks can yield better performance than English. However, this phenomenon remains under-explored.
In this paper, we explore the upper bound of harnessing multilingualism in reasoning tasks, suggesting that multilingual reasoning promises significantly~(by nearly 10 Acc@$k$ points) and robustly~(tolerance for variations in translation quality and language choice) higher upper bounds than English-only reasoning. Besides analyzing the reason behind the upper bound and challenges in reaching it, we also find that common answer selection methods cannot achieve this upper bound, due to their limitations and biases. These insights could pave the way for future research aimed at fully harnessing the potential of multilingual reasoning in LLMs~\footnote{Our code is available at \url{https://github.com/CONE-MT/multilingual_reasoning}}.
\end{abstract}

\section{Introduction}

Leading Large Language Models~(LLMs; \citealp{openai2023gpt4,team2024gemini,deepseek2025r1,wei2022chain,yao2023tree,li2025codeio}) are built on a robust multilingual foundation, developed through extensive exposure to diverse multilingual data~\cite{openai2023gpt4,gpt,palm,llama1,llama2} and effective vocabulary sharing across various languages~\cite{yuan-etal-2024-vocabulary}. However, due to the dominance of English in the training resources, the models exhibit a notable bias toward English; specifically, ~\textit{these models tend to achieve higher performance when tasks are presented in English}~\citep{huang-etal-2022-zero,fu-etal-2022-polyglot}, especially in reasoning-demanding tasks that are of great research interest~\cite{shi2023language,she-etal-2024-mapo,mindmerger,etxaniz-etal-2024-multilingual}.

\begin{figure}[t]
    \centering
    \includegraphics[width=\linewidth]{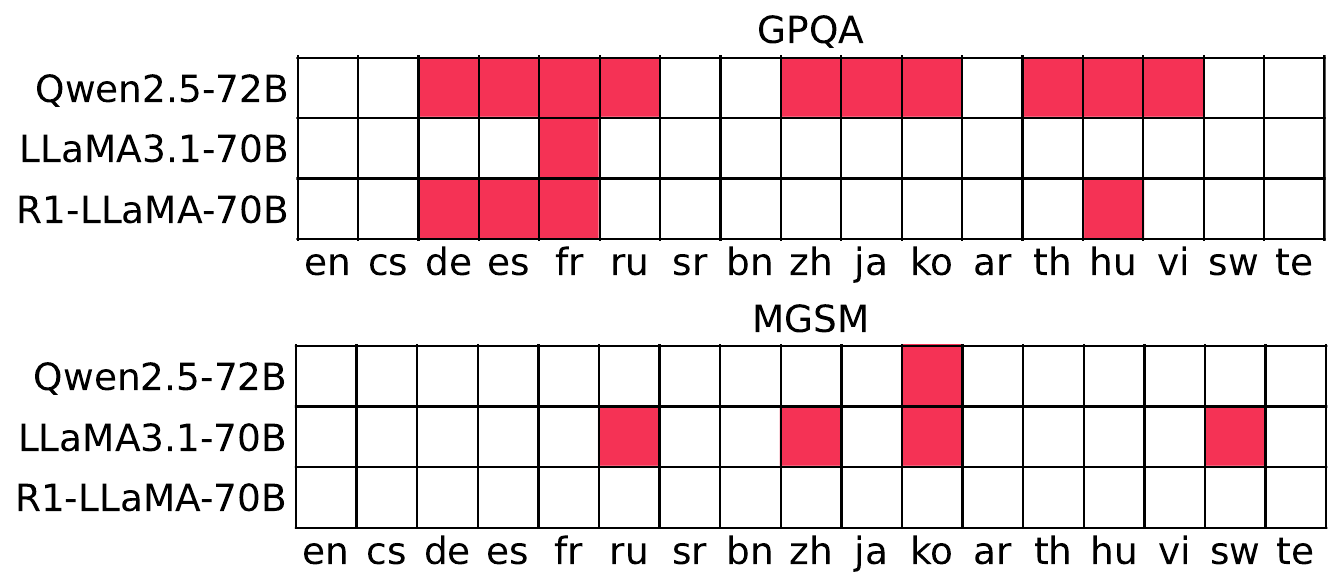}
    \caption{English is not always better than other languages. Evaluation results on the human-translated GPQA~\cite{rein2023gpqa} and MGSM~\cite{shi2023language}  datasets~(obtained from~\citet{huang2025benchmax}). The red cells indicate greater-than-English scores.}
    \label{fig:excel_english}
\end{figure}

As a result, multiple previous studies~\cite{lai-nissim-2024-mcot,zhu-etal-2024-question,she-etal-2024-mapo} aimed to improve non-English reasoning performance by aligning to English behaviors. However, as illustrated in Figure \ref{fig:excel_english}, recent LLMs sometimes show stronger reasoning ability in non-English~(not limited to high-resource ones) than in English~\cite{huang2025benchmax}, underlining the potential of thinking beyond English.
Inspired by this, we want to quantify the potential of multilingual thinking, compare it to English thinking, and explore the factors affecting its potential.

\textbf{So, how can we quantify the potential gain from multilingual thinking?} In this paper, we aggregate model responses to translated parallel inputs on two typical reasoning-specific tasks, namely GPQA~\cite{rein2023gpqa} and MGSM~\cite{shi2023language}, and evaluate the performance upper-bound of multilingual reasoning, calculated by Acc@$k$ (existence probability of at least one correct out of $k$ answers) on LLaMA3.1-70B, Qwen2.5-72B and R1-distill-LLaMA3.1-70B. 
The results show multilingual thinking can ideally boost GPQA accuracy from $\sim$45 to $\sim$90, and MGSM from $\sim$90 to $\sim$100.

\textbf{Is the gain genuinely a result of multilingual thinking, rather than from variations in input or randomness in decoding?} Compared to two English baselines: multiple random runs and paraphrased inputs, multilingual thinking starts to show advantages when using only 4 candidates, and the advantages continue to increase as more candidates are included. When using 17 languages (the total number of languages in \citeauthor{huang2025benchmax}), the gain from multilingual thinking surpasses the baselines by nearly 10 Acc@$k$ points. 


\textbf{What factors affects the multilingual thinking potential?} We conduct experiments regarding language selection, multilingual text quality and answer selection.
Previous experiments have shown that reasoning tasks can yield significant gains by aggregating results from 3-4 languages.  In our comprehensive tests, we randomly selected results from 4 out of 17 languages, and interestingly, the average performance was similar to that of the optimal combinations, demonstrating robustness in language selection. Furthermore, switching from machine translation to high-quality human translations does not lead to any substantial change in performance. However, when we shift Acc@$k$ to majority voting, the advantages of multilingual thinking diminish. Therefore, for multilingual thinking, answer selection is crucial to realize its benefits.

In addition to majority voting, other common answer selection strategies, prompt-based selection and LLM-as-a-judge selection~\cite{zhengJudgingLLMasajudgeMTbench2023a} still fail to elicit the potential of multilingual reasoning. Prompt-based selection tries to inject language-related guidance into instruction, such as predefined allowable languages or requirements of question translation, and LLM-as-a-judge attempts to generate a multilingual response independently, but the final answer requires additional judgment from the LLM. Unfortunately, experimental results reveal that performance gains occur inconsistently across different settings, suggesting that a stable selection method for leveraging multilingualism for enhanced reasoning remains elusive.

\textbf{Finally, we analyze the possible reasons and challenges associated with multilingual thinking.} Our experimental results suggest that questions of varying difficulty have different language requirements, and some languages can compensate for errors made in others, performing correctly even when other languages do not. Meanwhile, the challenges of different answer selection methods are different. Majority voting performance does not necessarily improve with the number of languages used and it depends on the optimal combination of languages, which contrast with the trend observed in Acc@$k$. Additionally, prompt-based methods and LLM-as-a-judge selection can introduce language bias. The main contribution can be summarized as:

\begin{itemize} [nosep,itemsep=2pt,leftmargin=0.2cm]
    \item We comprehensively analyze how multilingualism can enhance reasoning performance, laying the groundwork for understanding its huge potential.
    \item We evaluate common answer selection methods and find it is a challenge to tap into the advantage of multilingualism, highlighting the difficulties.
    \item We analyze the potential reasons for multilingual thinking gain, point out the limitations of existing methods, and share interesting findings for future research.
\end{itemize}

\section{Related Work}
\paragraph{Enhancing LLMs' Reasoning Performance}
Enhancing reasoning capabilities has emerged as a central challenge in LLM research. 
Prior work has approached this challenge from both training and inference perspectives.

\begin{figure*}[h!]
    \centering
    \includegraphics[width=0.8\textwidth]{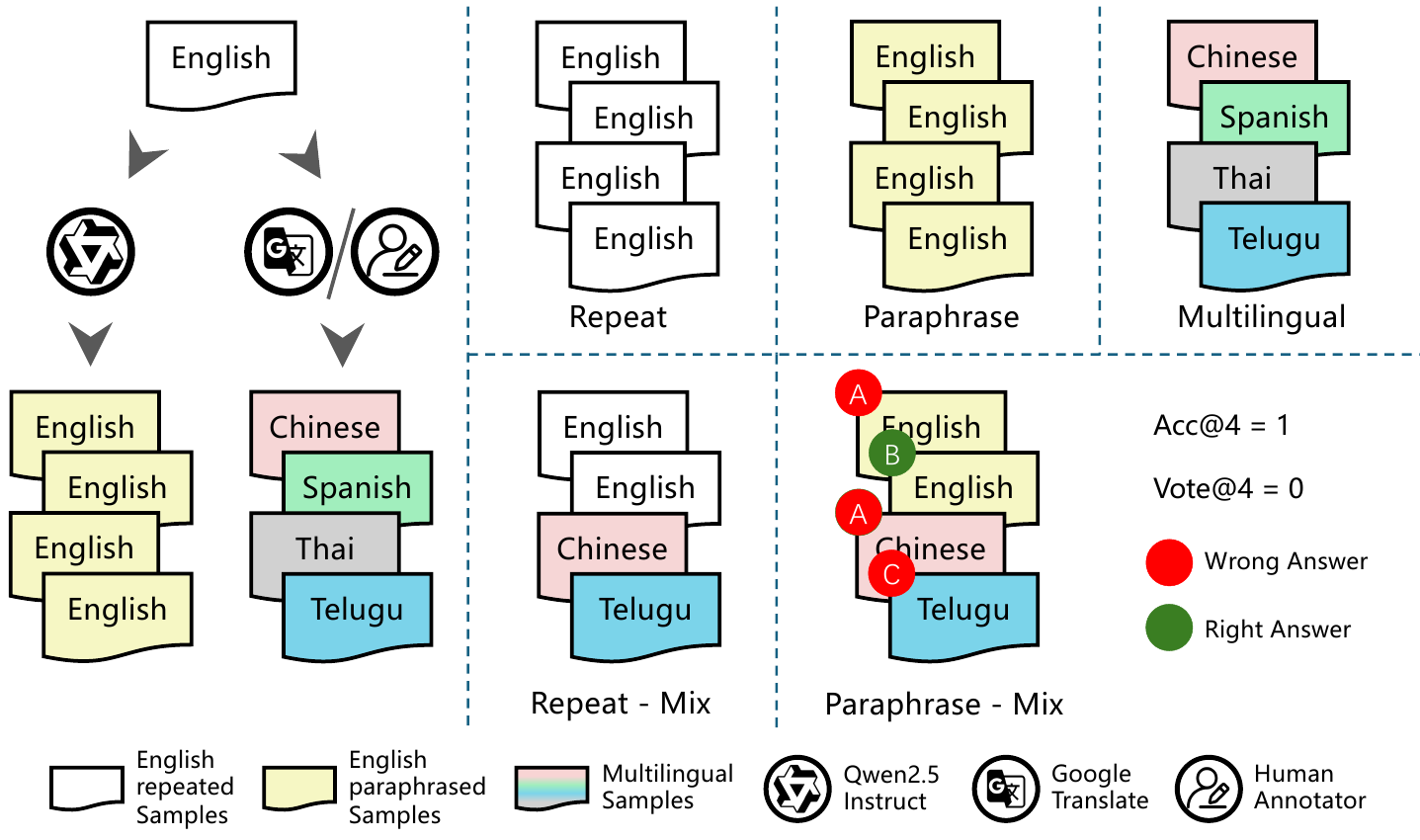}
    \caption{An introduction to input samples across various comparison methods, including Multilingual, Repeat, Paraphrase, Repeat-Mix, and Paraphrase-Mix.}
    \label{fig:method_overview}
\end{figure*}

The training perspective includes pretraining and post-training. For pretraining, \citet{taylor2022galacticalargelanguagemodel} used a specially designed training processes to optimize the model's STEM-related performance, and \citet{aryabumi2024code} explored the relationship between code pretraining and reasoning capabilities. Meanwhile, post-training approaches have also shown promising results in enhancing reasoning performance. \citet{hao-etal-2023-reasoning} proposed using Markov Decision Processes to model reasoning steps and using Monte-Carlo Tree Search to optimize the output; \citet{deepseek2025r1} proposed a novel training pipeline to elicit spontaneous long-chain reasoning in LLMs; And \citet{muennighoff2025s1} adopted small-scale instruction-tuning to boost the reasoning performance on competition math problems.

The inference perspective includes various prompting and sampling techniques.
In terms of prompting, chain-of-thought (CoT)~\cite{wei2022chain} and its variants \cite{yao2023tree,Besta2024graph} let models to break down complex problems into intermediate steps and achieve higher reasoning accuracy. Based on them, specialized CoT structures have been proposed for different tasks, including planning-oriented \cite{yasunagaLargeLanguageModels2023a,jiang2024selfplanningcodegenerationlarge,wang-etal-2023-plan,wangHypothesisSearchInductive2023} and symbolization-oriented \cite{gaur-saunshi-2023-reasoning,wang2023exploringequationbetterintermediate,chen2023program,li2023structuredchainofthoughtpromptingcode,xu-etal-2024-faithful,imani-etal-2023-mathprompter}. In terms of sampling, the self-consistency methods \citet{wangSelfConsistencyImprovesChain2022,wang-etal-2024-self-consistency,li-etal-2023-making} have been widely acknowledged. Our work pushes these studies further by focusing on the impact of multilingualism on LLM's reasoning behavior.

\paragraph{Multilingual Reasoning of LLMs}
Multilingual capability is crucial in LLM development. 
Earlier LLMs exhibited unbalanced performance across languages, with non-English CoT reasoning typically underperforming English CoT~\cite{shi2023language,qin-etal-2023-cross,she-etal-2024-mapo,mindmerger,etxaniz-etal-2024-multilingual}. To mitigate this gap, previous studies have mainly proposed multilingual CoT \cite{lai-nissim-2024-mcot,chai2024xcotcrosslingualinstructiontuning,huang-etal-2023-languages}, finetuning with translation data \cite{chen-etal-2024-breaking,zhu-etal-2024-question}, and multilingual preference training \cite{she-etal-2024-mapo,yangLanguageImbalanceDriven2024a}. However, recent advances in pre-trained language models have significantly transformed this landscape. 
Notably, \citet{huang2025benchmax} demonstrated that state-of-the-art LLMs such as \texttt{Qwen}~\cite{qwen2025qwen25} and \texttt{LLaMA}~\cite{dubey2024llama} achieve superior reasoning accuracy with non-English CoT compared to their English counterparts.
In this paper, we systematically investigate this phenomenon and explore how to leverage multilingual reasoning to probe LLMs' performance ceiling.

\section{High Upper Bound of Multilingual Reasoning}
\label{sec:few_huge_gain}

\subsection{Study Setup} 
\label{sec:setting}
To examine multilingual reasoning benefits, we analyze LLM responses to questions translated into multiple languages, and ablate the gains of increasing multilingualism versus increasing sampled response numbers (Figure~\ref{fig:method_overview}).
Specifically, we compare following approaches to transform question and collect LLM responses:
\begin{itemize}[nosep,itemsep=3pt,leftmargin=0.4cm]
    \item \textit{Multilingual}: The English samples are translated into various languages and then all fed into the model using a fixed random seed.
    \item \textit{Repeat}: The English samples are repeatedly fed into the model with different random seeds.
    \item \textit{Paraphrase}: The English samples are paraphrased by LLM and then all fed into the model using a fixed random seed.
    \item \textit{Repeat-Mix}: We combine Repeat and Multilingual samples in a $50/50$ proportion. 
    2 out of 4 random seeds are used by Repeat and 2 out of 17 languages are used by Multilingual.
    \item \textit{Paraphrase-Mix}: We mixed Paraphrase and Multilingual samples also in a $50/50$ proportion.
\end{itemize}

\paragraph{Models} We use Qwen2.5-72B~\cite{qwen2025qwen25}~\footnote{\url{Qwen/Qwen2.5-72B-Instruct}}, LLaMA3.1-70B~\cite{dubey2024llama}\footnote{\url{https://huggingface.co/meta-llama/Llama-3.1-70B-Instruct}} and R1-Distill-LLaMA-70B~\cite{deepseek2025r1}\footnote{\url{https://huggingface.co/deepseek-ai/DeepSeek-R1-Distill-Llama-70B}} in our experiments.
All results are based on their post-trained / instruction-tuned versions.
We prompt these models to employ Chain-of-Thought~(CoT) reasoning for all questions during inference.
The prompt template is reported in Appendix~\ref{sec:appendix_models}.

\paragraph{Testing Scenario}
We analyze the \textsc{GPQA}\citep{rein2023gpqa} and \textsc{MGSM}\citep{shi2023language} datasets, which are supported in 17 languages\footnote{English, Spanish, French, German, Russian, Bengali, Japanese, Thai, Swahili, Chinese, Telugu, Arabic, Korean, Serbian, Czech, Hungarian, and Vietnamese.} by BenchMAX~\citep{huang2025benchmax} with human translation. GPQA evaluates LLMs' reasoning from a scientific perspective, while MGSM assesses their mathematical reasoning capabilities. Additionally, we utilize Google Translate to provide machine translation in these languages and compare machine translation with human translation. 

\begin{figure}[!t]
    \centering
    \includegraphics[width=0.48\textwidth]{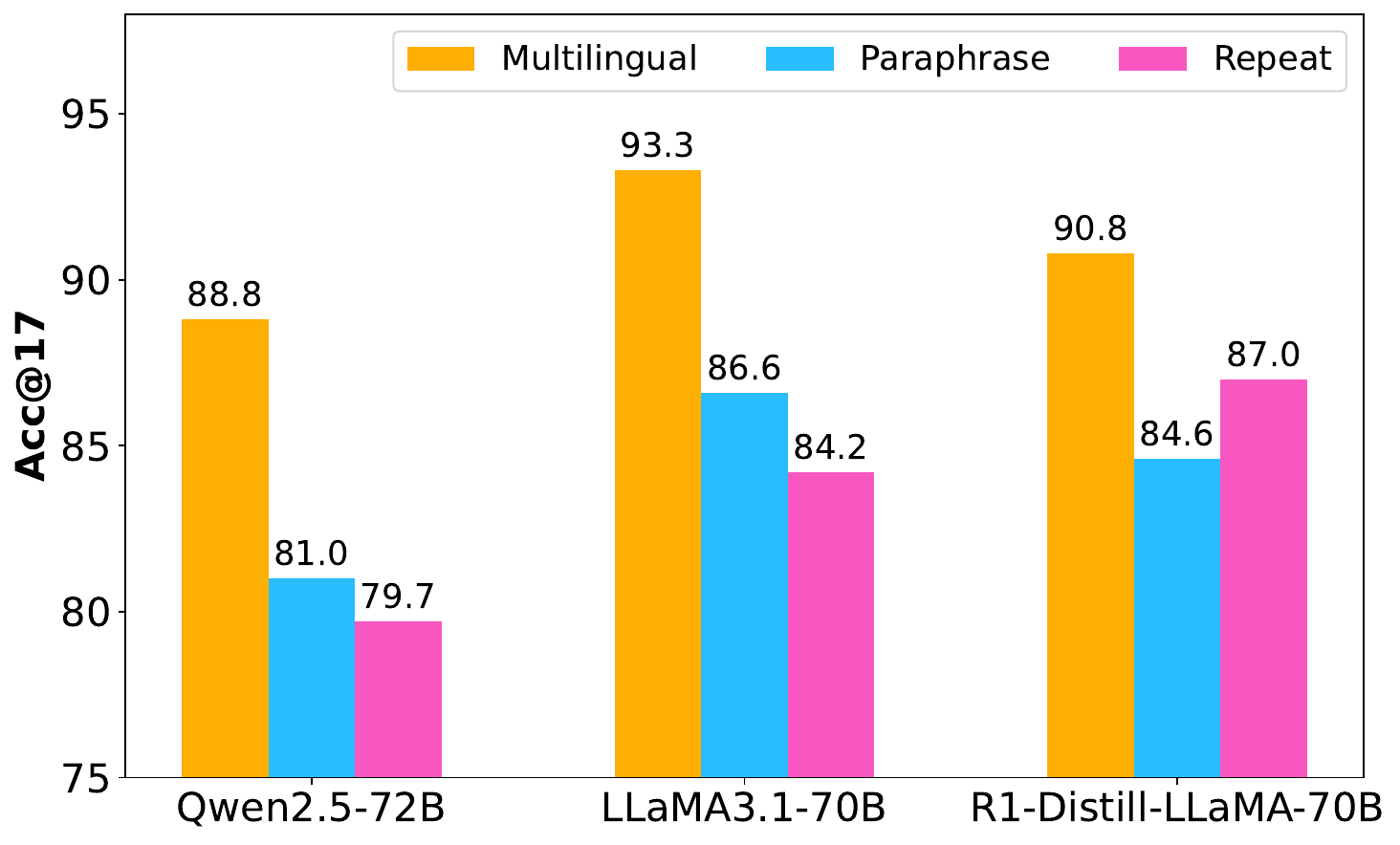}
    \caption{Compared to Repeat and Paraphrase, Multilingual demonstrates a higher performance upper bound. Acc@17 scores of Multilingual, Paraphrase and Repeat settings of the three models on the human-translated GPQA dataset.}
    \label{fig:higher_upper_bound}
\end{figure}

\paragraph{Metric} The default metric that we used is Accuracy~(Acc), which measures the agreement of the prediction generated by the model with the ground truth. 
$\overline{\text{Acc}}$ represents the average accuracy across $k$ answer candidates. 
We use Acc@$k$ metric to test the probability that at least one generated answer out of $k$ for a problem is the ground truth. 
Vote@$k$ is utilized to assess model's accuracy after selecting answers from $k$ candidates using a majority voting strategy. Judge@$k$ is used in the LLM-as-a-judge experiments to denote the accuracy of the judged winners.


\subsection{Intriguing Phenomena}
\label{sec:phenomena}

We notice that using multiple languages in reasoning is beneficial and results in significant performance improvements.


\paragraph{Phenomenon 1: Mixing languages boosts performance, setting higher upper bound.}
Enhancing language diversity during model generation results in remarkable performance improvements, with the ceiling of these improvements notably high. Compared to Repeat and Paraphrase, as depicted in Figure~\ref{fig:higher_upper_bound}, Multilingual can yield gains of around 8 accuracy points when using 17 candidates each. 


\paragraph{Phenomenon 2: A few languages offer substantial performance boosts.}

\begin{figure}[!t]
    \centering
    \includegraphics[width=0.48\textwidth]{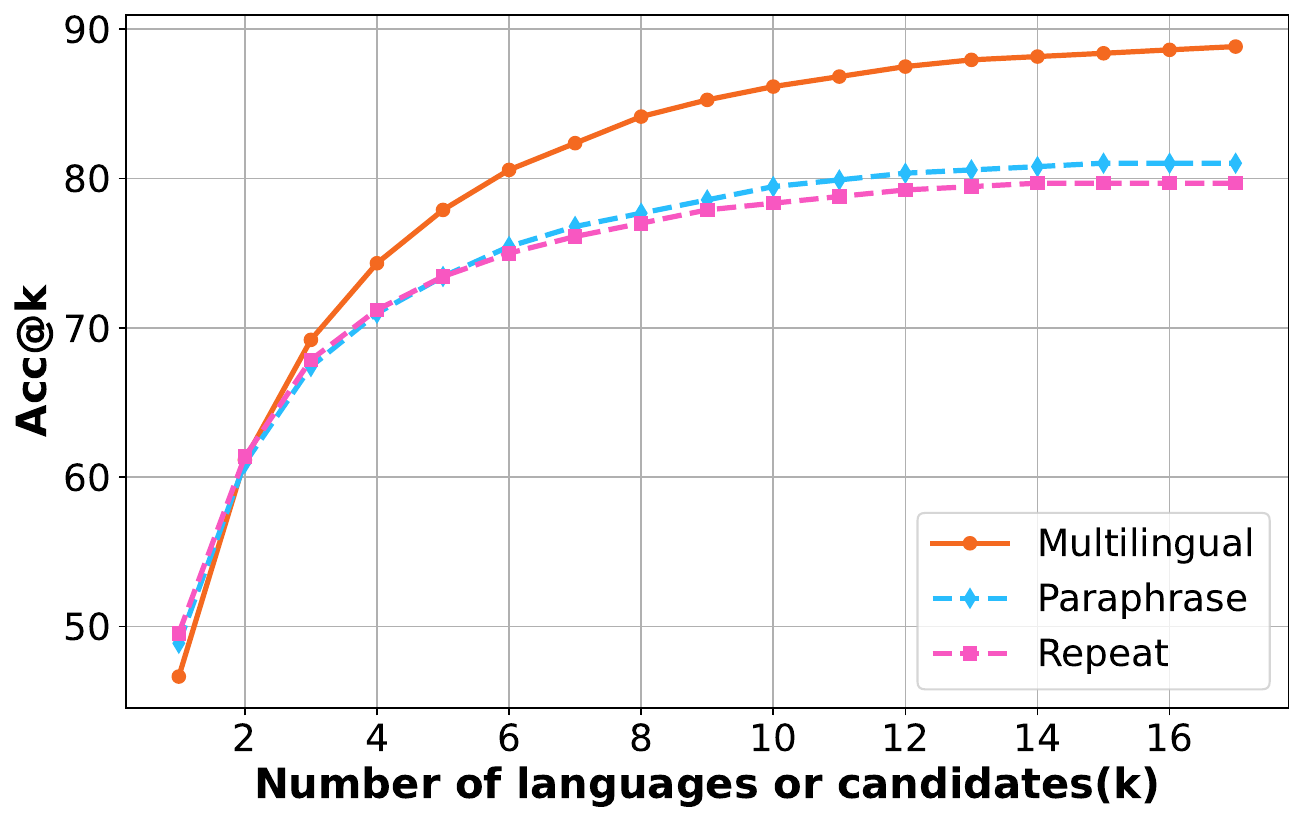}
    \caption{Multilingual surpasses Paraphrase and Repeat in Acc@$k$ after $k=3$ in a growing margin. Best Acc@$k$ (out of 17) of Multilingual, Paraphrase and Repeat settings for Qwen2.5-72B with increasing numbers of languages or candidates on the human-translated GPQA dataset.}
    \label{fig:xgpqa-n-langs}
\end{figure}

In GPQA task, we rank 17 languages based on their performance in each model from high to low and combined the top-performing languages with varying numbers each time. As shown in Figure~\ref{fig:xgpqa-n-langs}, as the number of candidate languages increases, the Acc@$k$ performance consistently improves. Notably, just a few ($\geqslant4$) languages can significantly enhance performance, quickly surpassing that of Repeat / Paraphrase.
Notably, even if none of the non-English language in the combination outperforms English, their combination can still achieve significant improvements.

\paragraph{Phenomenon 3: Multilingual gain - Going beyond existing English benefits}

\begin{figure}[!t]
    \centering
    \includegraphics[width=0.48\textwidth]{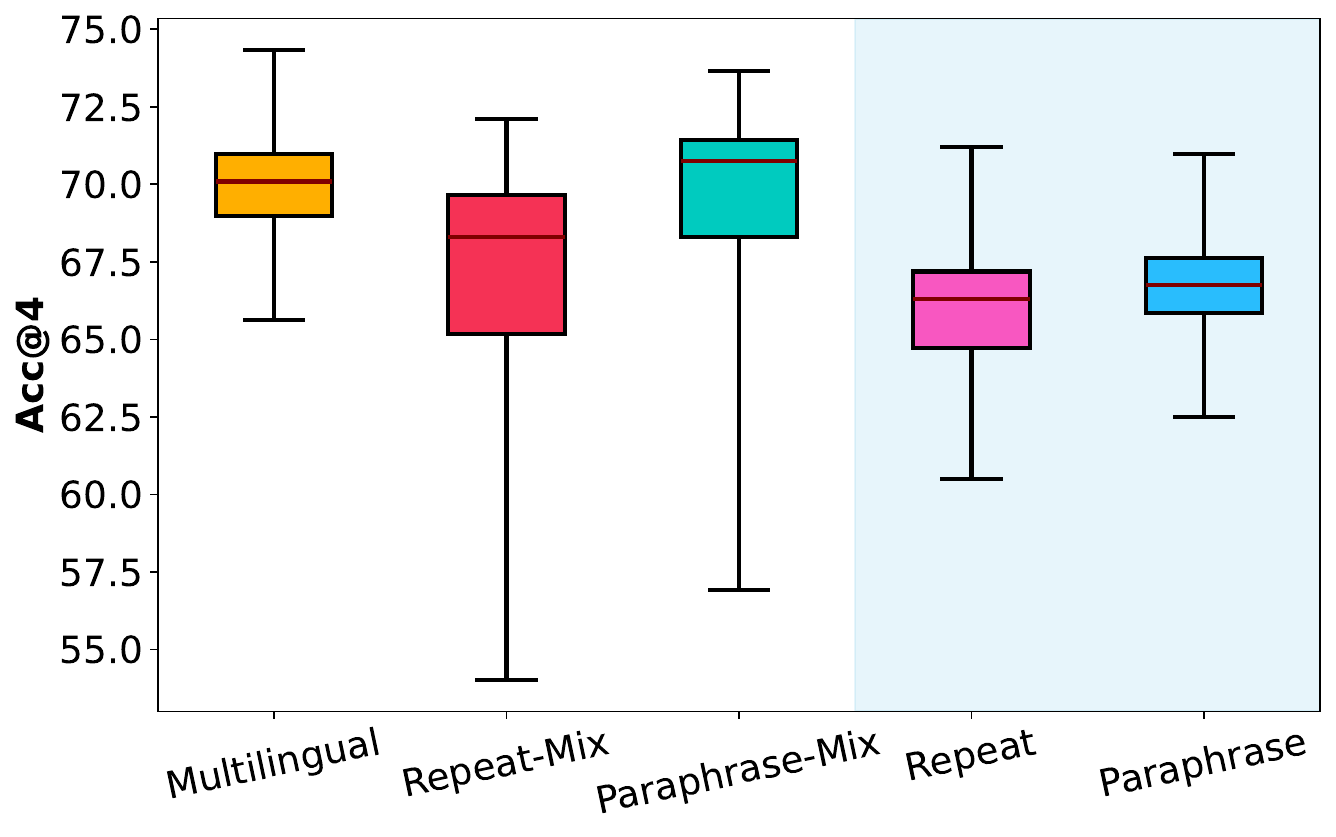}
    \caption{Fully utilizing non-English languages can improve the upper bound. Distribution of Acc@4 scores of all possible 4-candidate-combinations with Qwen2.5-72B on the human-translated GPQA dataset, under different settings.}
    \label{fig:multilingual_gain}
\end{figure}

As illustrated in Figure~\ref{fig:multilingual_gain}, Multilingual significantly improves reasoning performance, surpassing the limits achieved by Repeat or Paraphrase.
Notably, the improvements of Multilingual do not overlap with those derived from Repeat or Paraphrase. 
The experiments involving Repeat-Mix and Paraphrase-Mix indicate that replacing a portion of the input with multilingual data results in additional benefits to the performance upper bound. 
This suggests that Multilingual a unique advantage in reasoning tasks, enabling models to leverage diverse linguistic structures and contexts.

\paragraph{Phenomenon 4: The upper bound is tolerant of sub-optimal language choice.}


Table \ref{tab:lg_choice} shows the evaluation results of all 4-combinations out of the 17 testing languages, where "Best", "Worst" and "Random" indicate the highest, lowest and average scores among all the language combinations. Interestingly, while the mean single-language performance varies across combinations (indicated by $\overline{\text{Acc}}$), the "Random" Acc@$k$ scores are close to "Best", meaning a randomly selected language combination is expected to have similar upper bound to that of the best-performing combination.

\begin{table}[!h]
    \centering
    \footnotesize
    \begin{tabular}{c|c|cc}
    \toprule
    \textbf{Model} & \textbf{Combo} & $\overline{\textbf{Acc}}$ & \textbf{Acc@4} \\
    \midrule
         \multirow{3}{*}{Qwen2.5-72B} & Best & 43.7 & 74.3 \\
         & Worst & 37.8 & 65.6 \\
         & Random & 41.5 & 70.0 \\
    \midrule
         \multirow{3}{*}{LLaMA3.1-70B} & Best & 38.0 &	73.9\\
         & Worst & 32.6 & 65.2 \\
         & Random & 36.9 & 70.2 \\
    \midrule
        \multirow{3}{*}{R1-Distill-LLaMA-70B} & Best & 51.6 & 80.1 \\
        & Worst & 34.0 & 64.7 \\
        & Random & 49.0 & 75.5 \\
    \bottomrule
    \end{tabular}
    \caption{Multilingual upper bound is robust to language combination choices. Mean Acc ($\overline{\textbf{Acc}}$) and \textbf{Acc@4} of the best, worst and random language combinations (\textbf{Combo}) with the Multilingual setting on the human-translated GPQA dataset. While $\overline{\text{Acc}}$ varies, the gain in Acc@$k$ remains high.}
    \label{tab:lg_choice}
\end{table}



\begin{figure}[!t]
    \centering
    \includegraphics[width=0.48\textwidth]{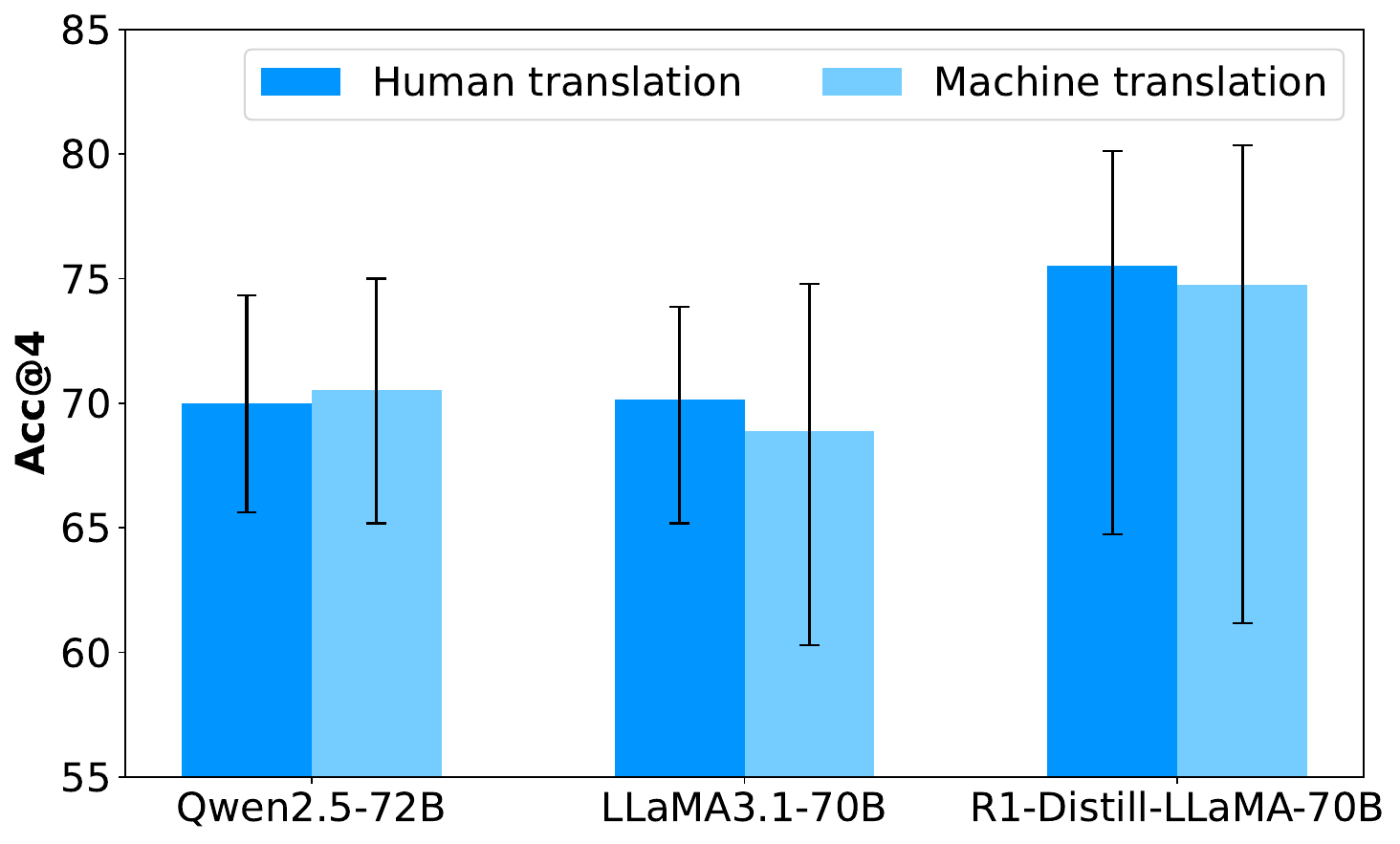}
    \caption{The Multilingual upper bound is stable regardless of the question translation quality. Comparison of Acc@4 on human- and machine-translated GPQA dataset among all possible 4-language combinations in Multilingual setting. The values and error bars denote mean, max and min scores.}
    \label{fig:translation_quality}
\end{figure}

\paragraph{Phenomenon 5: The upper bound is robust to translation quality.}
Obtaining a human-translated multilingual dataset of high quality is challenging and is not scalable for many tasks. Therefore, we investigate whether human translation quality is essential for model performance. As depicted in Figure~\ref{fig:translation_quality}, the models exhibit slight differences in performances of random and best language combinations, when tested on human-translated and machine-translated datasets. This experiment highlights that we can elicit multilingual reasoning with machine-translated data to upper bounds as high as with human-translated data.

\section{Low Performance of Common Answer Selection Strategies}
\label{sec:selection-strategy}
In this section, we explore commonly methods to reach the upper bound, including  Majority voting~(\S\ref{sec:majority_voting}), Prompt-based language selection~(\S\ref{sec:prompt_selection}), and LLM-as-a-judge selection~(\S\ref{sec:LLM_as_a_judge}).

\subsection{Majority voting}
\label{sec:majority_voting}
We conduct a series of experiments across different models to evaluate the effectiveness of various answer selection strategies, specifically Repeat, Paraphrase, and Multilingual settings, as shown in Table~\ref{tab:answer-selection_methods}. In the Acc@4 metric, the Multilingual approach outperforms the other strategies. However, when analyzing the results with Vote@4, Multilingual does not achieve similarly favorable outcomes. This inconsistency suggests that while Multilingual strategies can be advantageous in certain contexts, their overall effectiveness may be limited by the answer selection criteria employed. These findings underscore the necessity of different answer selection strategies across various models.

\begin{table}[!h]
\centering
\resizebox{\linewidth}{!}{
\begin{tabular}{c|ccccc}
\toprule
\multirow{2}{*}{\textbf{Model}} & \multirow{2}{*}{\textbf{Setting}} & \multicolumn{2}{c}{\textbf{Best Combo}} & \multicolumn{2}{c}{\textbf{Random Combo}} \\
 &  & \textbf{Acc@4} & \textbf{Vote@4} & \textbf{Acc@4} & \textbf{Vote@4} \\ \midrule
\multirow{4}{*}{\begin{tabular}[c]{@{}c@{}}Qwen2.5-\\ 72B\end{tabular}} & Repeat & 71.2 & 53.7 & 65.9 & \textbf{53.6} \\
 & Paraphrase & 71.0 & \textbf{54.4} & 66.7 & 52.8 \\
 & Multilingual-h & 74.3 & 54.2 & 70.0 & 51.7 \\
 & \multicolumn{1}{l}{Multilingual-m} & \textbf{75.0} & 53.0 & \textbf{70.5} & 52.2 \\ \midrule
\multirow{4}{*}{\begin{tabular}[c]{@{}c@{}}LLaMA3.1-\\ 70B\end{tabular}} & Repeat & 71.0 & 50.4 & 66.4 & 49.8 \\
 & Paraphrase & 73.0 & 51.3 & 68.7 & \textbf{51.6} \\
 & Multilingual-h & 73.9 & 49.8 & \textbf{70.2} & 48.8 \\
 & \multicolumn{1}{l}{Multilingual-m} & \textbf{74.8} & \textbf{54.4} & 68.9 & 49.0 \\ \midrule
\multirow{4}{*}{\begin{tabular}[c]{@{}c@{}}R1-Distill-\\ LLaMA-70B\end{tabular}} & Repeat & 77.9 & 63.2 & 74.3 & \textbf{62.9} \\
 & Paraphrase & 74.8 & 60.8 & 71.1 & 59.9 \\
 & Multilingual-h & 80.1 & 61.2 & \textbf{75.5} & 60.0 \\
 & \multicolumn{1}{l}{Multilingual-m} & \textbf{80.4} & \textbf{64.6} & 74.7 & 60.1 \\ \bottomrule
\end{tabular}
}
\caption{Answer selection is challenging and critical for effective Multilingual reasoning. Comparison of Acc@4 and Vote@4 with the best (\textbf{Best Combo}) and random 4-combinations (\textbf{Random Combo}) out of 17 languages/candidates with respect to Acc@4, for each model under different settings on the human-translated (Multilingual-h) and machine-translated (Multilingual-m) GPQA datasets.}
\label{tab:answer-selection_methods}
\end{table}

\subsection{Prompt-based language selection}
\label{sec:prompt_selection}

One straightforward method for answer selection is the prompt-based approach, which entails furnishing precise input instructions to direct the model in producing the desired outputs. To steer the model towards maximizing its multilingual capabilities, we have customized prompts for the LLM from three crucial viewpoints: language constraint, English allowance, and question translation.

\begin{itemize}[nosep,itemsep=2pt,leftmargin=0.3cm]
    \item Language Constraint~(LC) means whether to provide a predefined set of languages that the model can utilize.
    \item English Allowance~(EA) means to whether to incorporate English as one of the languages that can be used.\footnote{For LC and EA, we use the best language combinations for each model observed on the human-translated GPQA dataset, and substitute between English and non-English candidates while keeping the upper-bound performances as close as possible.}
    \item Question Translation~(QT) means whether to explicitly prompt the model to translate the questions to the target languages thus encouraging multilingual responses.
\end{itemize}

\begingroup

\begin{table}[!h]
    \centering
    \scriptsize
    \resizebox{0.95\linewidth}{!}{
    \begin{tabular}{c|cccc|cc}
    \toprule
         \textbf{Model} & \textbf{LC} & \textbf{EA} & \textbf{QT} & \textbf{Setting} & \textbf{Acc@4} & \textbf{Vote@4}\\
         \midrule
         \multirow{8}{*}{\makecell{Qwen2.5-\\72B}} & - & - & - & Repeat &  65.9 & \textbf{53.6}\\
         & - & - & - & Paraphrase &  \textbf{66.7} & 52.8 \\
          & \Checkmark & \XSolidBrush & \Checkmark & - & 59.2 & 48.2 \\
         & \Checkmark & \Checkmark & \Checkmark &  - & 63.8 & 51.8 \\
         & \XSolidBrush & \Checkmark & \Checkmark & - & 61.2 & 53.2 \\
         & \Checkmark & \XSolidBrush & \XSolidBrush & - & 62.7 & 50.6 \\
         & \Checkmark & \Checkmark & \XSolidBrush & - & 61.2 & 52.5 \\
         & \XSolidBrush & \Checkmark & \XSolidBrush & - &62.1 & 52.0 \\
         \midrule
         
        \multirow{8}{*}{\makecell{LLaMA3.1-\\70B}} & - & - & - & Repeat &  66.4 & 49.8 \\
         & - & - & - & Paraphrase &  \textbf{68.7} & \textbf{51.6} \\
         & \Checkmark & \XSolidBrush & \Checkmark &  - & 58.9 & 46.6 \\
         & \Checkmark & \Checkmark & \Checkmark &  - & 61.8 & 47.5 \\
         & \XSolidBrush & \Checkmark & \Checkmark &  - & 65.6 & 50.1 \\
         & \Checkmark & \XSolidBrush & \XSolidBrush &  - & 62.5 & 46.6 \\
         & \Checkmark & \Checkmark & \XSolidBrush &  - & 63.2 & 50.6 \\
         & \XSolidBrush & \Checkmark & \XSolidBrush &  - & 65.0 & 49.6 \\
         \midrule
         
        \multirow{8}{*}{\makecell{R1-Distill-\\LLaMA-70B}} & - & - & - & Repeat &   74.3 & 62.9\\
         & - & - & - & Paraphrase &  71.1 & 59.9 \\
        & \Checkmark & \XSolidBrush & \Checkmark &  - & 75.9 & 64.9 \\
         & \Checkmark & \Checkmark & \Checkmark &  - & 73.2 & 59.2 \\
         & \XSolidBrush & \Checkmark & \Checkmark &  - & 72.8 & 58.7 \\
         & \Checkmark & \XSolidBrush & \XSolidBrush &  - & \textbf{76.8} & \textbf{66.3} \\
         & \Checkmark & \Checkmark & \XSolidBrush &  - & 72.8 & 56.8 \\
         & \XSolidBrush & \Checkmark & \XSolidBrush &  - & 72.8 & 57.7 \\
         \bottomrule
    \end{tabular}}
    \caption{Different prompt-based settings show little performance difference, and self-translation is not the key setting. Acc@4 and Vote@4 of prompt-based selection methods, compared with the random-4 performances of Repeat and Paraphrase on the English GPQA dataset. LC, EA and QT stand for Language Constraint, English Allowance and Question Translation.}
    \label{tab:prompt_based}
\end{table}
\endgroup

Prompt-based methods cannot unlock a model's multilingual capabilities during reasoning. 
As shown in Table~\ref{tab:prompt_based}, no prompt-based approach stands out as superior, with minimal differences between methods.  However, there are still some intriguing discoveries:1) Translating the original question from English to non-English before responding does not affect the model's final performance. 
2) Interestingly, when comparing LLaMA3.1-70B with R1-Distill-Llama-70B, prompt-based methods achieve better results than Repeat and Paraphrase.

\subsection{LLM-as-a-judge selection}
\label{sec:LLM_as_a_judge}
Another commonly used answer selection method is LLM-as-a-judge~\cite{li2024crowd}, where a judge model evaluates two answers to a given question, and selects the best of them as the winner.
Here, we use the tested models to judge their own outputs, and conduct pairwise judgments for each two of the candidates with position swapping, and take the one winning the most battles.

To test the effectiveness of this method, we run judges on the machine-translated multilingual questions, using the best language combination found on them for each model and collect the accuracies of the judged outputs (Judge@$k$). Then, we compare them with English Repeat and Paraphrase with the same judging process.

The results are shown in Table \ref{tab:llm-judge-xgpqa}. Still, while Multilingual leads the Acc@$k$ scores, its Judge@$k$ scores are lower than the English baselines except for the R1-Distill-LLaMA model. 
Also, the Judge@$k$ scores in most of the tested settings are lower than Vote@$k$ scores, meaning LLM-as-a-judge is even less effective than simple majority voting in answer selection. This suggests LLM-as-a-judge answer selection is not satisfactory.

\begin{table}[!t]
\centering
\scriptsize
\begin{tabular}{c|cccc}
\toprule
\textbf{Model} & \textbf{Setting} & \textbf{Acc@4*} & \textbf{Vote@4*} & \textbf{Judge@4*} \\ \hline
\multirow{4}{*}{\begin{tabular}[c]{@{}c@{}}Qwen2.5-\\ 72B\end{tabular}} & Repeat & 61.4 & 53.4 & 48.9 \\
 & Paraphrase & 63.0 & \textbf{54.2} & \textbf{50.4} \\
 & Multilingual-h & \textbf{67.0} & 53.0 & 48.0 \\
 & Multilingual-m & 66.7 & 51.4 & 46.4 \\ \midrule
\multirow{4}{*}{\begin{tabular}[c]{@{}c@{}}LLaMA3.1-\\ 70B\end{tabular}} & Repeat & 62.1 & 50.6 & \textbf{47.1} \\
 & Paraphrase & 65.8 & 49.2 & 46.2 \\
 & Multilingual-h & 67.2 & 50.2 & 39.3 \\
 & Multilingual-m & \textbf{67.6} & \textbf{50.9} & 41.3 \\ \midrule
\multirow{4}{*}{\begin{tabular}[c]{@{}c@{}}R1-Distill-\\ LLaMA-70B\end{tabular}} & Repeat & 71.2 & 57.2 & 57.1 \\
 & Paraphrase & 71.9 & 59.2 & 58.9 \\
 & Multilingual-h & \textbf{76.6} & 62.3 & 60.7 \\ 
 & Multilingual-m & 76.1 & \textbf{62.6} & \textbf{62.3} \\ \bottomrule
\end{tabular}
\caption{LLM-as-a-judge exhibits Multilingual advantage only with R1-Distill-LLaMA-70B, which is not satisfactory. LLM-as-a-judge performance on the human-translated~(Multilingual-h) and machine-translated~(Multilingual-m) GPQA datasets. The asterisks(*) indicate that we only include 4 runs in each setting, using the best language combination for the dataset due to the cost of LLM judging, so the results are different from those in the previous tables.}
\label{tab:llm-judge-xgpqa}
\end{table}

\section{Analysis}
While \S\ref{sec:few_huge_gain} shows the high upper bound gain of multilingualism, \S\ref{sec:selection-strategy} shows that common selection approaches have difficulty realizing this gain. Here, we discuss the reasons behind this gap.

\subsection{Possible reasons for the upper bound gain}
We propose several possible reasons for the upper bound gain of multilingualism.

\paragraph{Language correctness correlates with question difficulty.}
The first hypothesis is that different languages match questions of different levels of difficulty. For questions in each level, there can be certain suitable languages for the models to use to achieve higher accuracies. 

To verify this hypothesis, on the GPQA tasks where question difficulties are labeled, we calculate the per-language accuracy on questions with different difficulty levels. The results (Table \ref{tab:difficulty-distribution}) show that the accuracy per language varies, and for each model, there are at least two different languages leading the accuracies on different difficulty levels. This indicates different languages indeed match different difficulty levels.

\begin{table}[!h]
\centering
\scriptsize
\resizebox{\linewidth}{!}{
\begin{tabular}{c|ccccc}
\toprule
\textbf{Model} & \textbf{\begin{tabular}[c]{@{}c@{}}Combo\\ Lang\end{tabular}} & \textbf{\begin{tabular}[c]{@{}c@{}}Easy\\ Undergrad\end{tabular}} & \textbf{\begin{tabular}[c]{@{}c@{}}Hard\\ Undergrad\end{tabular}} & \textbf{\begin{tabular}[c]{@{}c@{}}Hard\\ Grad\end{tabular}} & \textbf{\begin{tabular}[c]{@{}c@{}}Post-\\ Grad\end{tabular}} \\ \midrule
\multirow{4}{*}{\begin{tabular}[c]{@{}c@{}}Qwen2.5-\\ 72B\end{tabular}} & en & 47.6 & \textbf{50.6} & 41.2 & \textbf{44.1} \\
 & es & \textbf{57.1} & 44.6 & \textbf{43.1} & 38.2 \\
 & ja & 47.6 & 46.8 & 41.8 & 35.3 \\
 & th & 47.6 & 41.6 & 39.2 & 32.4 \\ \midrule
\multirow{4}{*}{\begin{tabular}[c]{@{}c@{}}LLaMA3.1-\\ 70B\end{tabular}} & fr & \textbf{47.6} & \textbf{42.5} & \textbf{41.8} & 35.3 \\
 & ko & 23.8 & 39.9 & 37.9 & \textbf{38.2} \\
 & sw & \textbf{47.6} & 31.8 & 30.1 & 35.3 \\
 & vi & \textbf{47.6} & 38.6 & 39.9 & \textbf{38.2} \\ \midrule
\multirow{4}{*}{\begin{tabular}[c]{@{}c@{}}R1-Distill-\\ LLaMA-70B\end{tabular}} & ar & 61.9 & 49.8 & 48.4 & 41.2 \\
 & es & 66.7 & \textbf{58.4} & \textbf{54.2} & \textbf{58.8} \\
 & ko & \textbf{76.2} & 54.9 & 47.7 & 50.0 \\
 & sr & 57.1 & 48.9 & 45.8 & 35.3 \\ \bottomrule
\end{tabular}
}
\caption{Languages to some extent match difficulty levels. Per-language accuracies across difficulty levels on the human-translated GPQA dataset, where the languages are from the best-performing combinations (\textbf{Combo Lang}). Each difficulty has one or more advantage languages.}
\label{tab:difficulty-distribution}
\end{table}





\paragraph{Existence of key advantage languages}
Another hypothesis is that, for a model on a specific task with multilingual reasoning, there will be some key advantage languages that often compensate for errors in other languages, which contributes to the high Acc@$k$. Furthermore, if the key advantage languages overlap in different models, it will be likely that these languages are more suitable than others on the specific task.

We set a standard called minority-majority overlap to identify such a language advantage. First, we collect the languages with high accuracies, both on questions correctly answered only in a few languages and by a vast majority of the languages.
Then, we report the overlap of the leading languages in the both situations. Finally, we report the cross-model overlap of these languages. As shown in Table \ref{tab:key-advantage-languages}, each model has some key advantage languages in the two tasks, respectively, and there are also cross-model key advantage languages, namely French for GPQA and Korean and English for MGSM.

\begin{table}[!t]
\centering
\scriptsize
\resizebox{0.95\linewidth}{!}{
\begin{tabular}{ccc}
\toprule
\textbf{Task}                      & \textbf{Model}                & \textbf{Advantage Langs}                      \\ \midrule
                          & Qwen2.5-72B          & ja,en,{\color[HTML]{34C724}fr},hu                           \\
                          & LLaMA3.1-70B         &  hu,en,{\color[HTML]{34C724}fr},ru,de \\
\multirow{-3}{*}{GPQA} & R1-Distill-LLaMA-70B & es,vi,cs,{\color[HTML]{34C724}fr}                           \\ \midrule
                          & Qwen2.5-72B          & {\color[HTML]{34C724}ko},ar,es,{\color[HTML]{34C724}en},sr,vi,hu                  \\
                          & LLaMA3.1-70B         & ru,{\color[HTML]{34C724}ko},{\color[HTML]{34C724}en},es,vi,de                     \\
\multirow{-3}{*}{MGSM}    & R1-Distill-LLaMA-70B & sr,ar,{\color[HTML]{34C724}ko},{\color[HTML]{34C724}en},cs,hu                     \\ \bottomrule
\end{tabular}}
\caption{Each model has key advantage languages that often compensate for errors in other languages in the two tasks, and there is cross-model overlap. Key advantage languages~(\textbf{Advantage Langs}) found by minority-majority overlap that filters language leading accuracy on questions correctly answered by a few or many tested languages.}
\label{tab:key-advantage-languages}
\end{table}

\subsection{Challenges to achieve the upper bound}
We will discuss some challenges in meeting the multilingual reasoning upper bound with common approaches.

\begin{figure}[!t]
    \centering
    \includegraphics[width=0.95\linewidth]{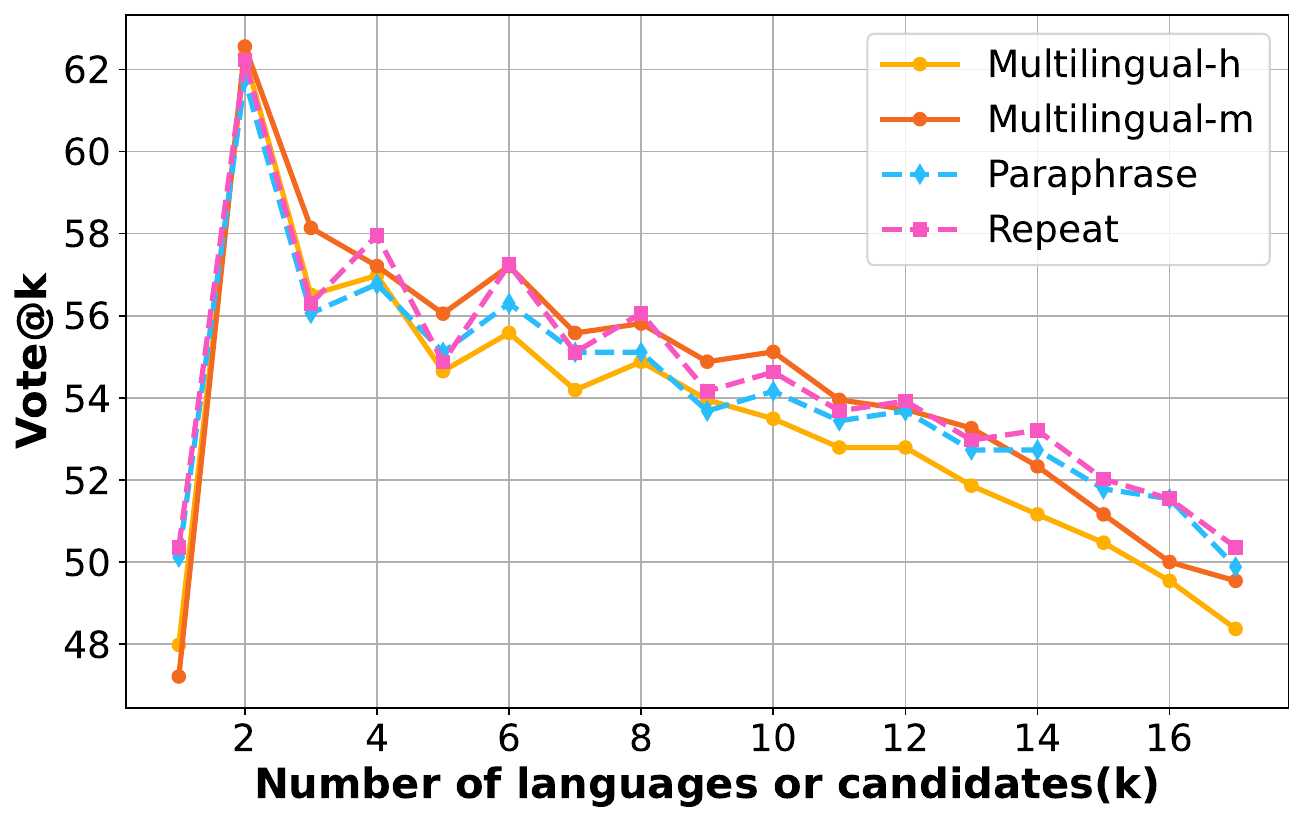}
    \caption{Voting performance does not increase with candidate number. Best Vote@$k$ (out of 17) of Paraphrase, Repeat and Multilingual with human (Multilingual-h) and machine translation (Multilingual-m) on the GPQA dataset for Qwen2.5-72B with increasing numbers of languages or candidates.}
    \label{fig:xgpqa-n-langs-vote}
\end{figure}

\paragraph{Voting performance does not grow with language numbers.}
As shown in Figure \ref{fig:xgpqa-n-langs-vote}, as the size of the language combination increases, the Vote@$k$ score does not increase but instead decreases, which is the opposite to the Acc@$k$ curve in Figure \ref{fig:xgpqa-n-langs}. This is mainly because the gain and advantage of multilingualism in Acc@$k$ is often brought by only a few languages, especially when the majority is wrong. Thus, a larger number of languages can bring more noise, making it harder for the correct answer to win majority.

\begin{figure}[!t]
    \centering
    \includegraphics[width=0.48\textwidth]{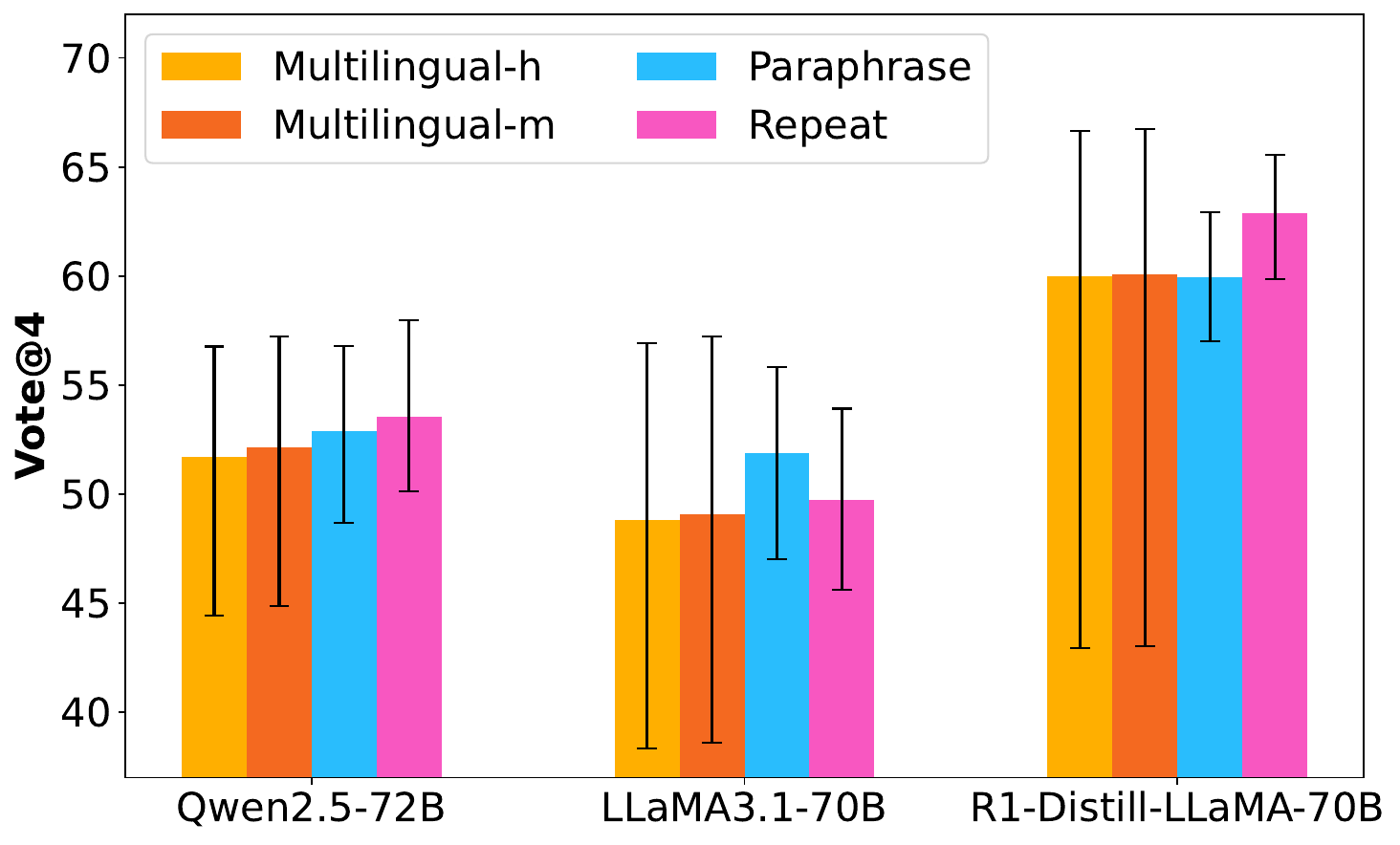}
    \caption{Multilingual performance is sensitive to the optimality of the language combination. Comparison of Vote@4 of Repeat, Paraphrase and Multilingual with human (Multilingual-h) and machine translation (Multilingual-m) on the GPQA dataset. The values and error bars denote mean, max and min scores.}
    \label{fig:xgpqa-vote-comparison}
\end{figure}

\paragraph{Voting performance relies on optimal language combination.}
While we show the multilingual reasoning upper bound is tolerant to sub-optimal language combinations in \S\ref{sec:phenomena}, the multilingual majority voting performance relies on optimal language combinations to surpass English voting. As shown in Figure \ref{fig:xgpqa-vote-comparison}, the voting accuracy of Multilingual is higher than or quite close to the those of Paraphrase and Repeat if all of them use their best language combinations. However, when the language combination is random or the worst, the Multilingual voting accuracy will be lower than the other two, indicating that majority voting on the Multilingual setting is sensitive to the optimality of the language combination.

\paragraph{Prompt-based and LLM-as-a-judge selection have language bias.}

For prompt-based selection, the models tend to choose high-resource languages for all the questions, thus decreasing the diversity of the Multilingual candidates. Table \ref{tab:promt-based-chosen-rate} shows the chosen rates of English and the most frequently chosen non-English language in different settings. The results show that, when English is allowed, the models will choose English in most cases; and when it is not allowed, the models tend to choose a certain language (such as Spanish or Vietnamese) than other languages in most cases.

\begin{table}[!t]
\centering
\scriptsize
\begin{tabular}{c|ccc|cc}
\toprule
\textbf{Model} & \textbf{LC}                 & \textbf{EA}                 & \textbf{QT}                 & \textbf{En} & \makecell{\textbf{Max}\\\textbf{Non-En}} \\ \midrule
\multirow{6}{*}{Qwen2.5-72B} &
  \Checkmark &
  \XSolidBrush &
  \Checkmark &
  4.4 &
  45.5 \\
               & \Checkmark   & \Checkmark   & \Checkmark   & 99.9     & 0.1              \\
               & \XSolidBrush & \Checkmark   & \Checkmark   & 99.7     & 0.3              \\
               & \Checkmark   & \XSolidBrush & \XSolidBrush & 62.1     & 17.2             \\
               & \Checkmark   & \Checkmark   & \XSolidBrush & 99.8     & 0.2              \\
               & \XSolidBrush & \Checkmark   & \XSolidBrush & 99.8     & 0.2              \\ \midrule
\multirow{6}{*}{LLaMA3.1-70B} &
  \Checkmark &
  \XSolidBrush &
  \Checkmark &
  1.1 &
  83.4 \\
               & \Checkmark   & \Checkmark   & \Checkmark   & 46.5     & 53.3             \\
               & \XSolidBrush & \Checkmark   & \Checkmark   & 99.7     & 0.2              \\
               & \Checkmark   & \XSolidBrush & \XSolidBrush & 25.6     & 52.5             \\
               & \Checkmark   & \Checkmark   & \XSolidBrush & 85.6     & 14.1             \\
               & \XSolidBrush & \Checkmark   & \XSolidBrush & 99.9     & 0.1              \\ \midrule
\multirow{6}{*}{R1-Distill-LLaMA-70B} &
  \Checkmark &
  \XSolidBrush &
  \Checkmark &
  100.0 &
  0.0 \\
               & \Checkmark   & \Checkmark   & \Checkmark   & 99.9     & 0.1              \\
               & \XSolidBrush & \Checkmark   & \Checkmark   & 99.9     & 0.1              \\
               & \Checkmark   & \XSolidBrush & \XSolidBrush & 99.9     & 0.1              \\
               & \Checkmark   & \Checkmark   & \XSolidBrush & 99.9     & 0.1              \\
               & \XSolidBrush & \Checkmark   & \XSolidBrush & 99.8     & 0.1              \\ \bottomrule
\end{tabular}
\caption{With prompt-based answer selection, models have a strong tendency to choose 1-2 certain languages instead of all others. Chosen rates of English and the highest non-English language, with the prompt-based answer selection methods.}
\label{tab:promt-based-chosen-rate}
\end{table}

\begin{table*}[!t]
\resizebox{\textwidth}{!}{
\centering
\scriptsize
\begin{tabular}{c|ccc||ccc}
\toprule
\multirow{2}{*}{\textbf{Model}}  & \multicolumn{3}{c||}{\textbf{Human-Translated Dataset}} & \multicolumn{3}{c}{\textbf{Machine-Translated Dataset}} \\
& \textbf{Lang} & \textbf{\begin{tabular}[c]{@{}c@{}}$\textbf{P}(\mathrm{\textbf{Chosen}}|\mathrm{\textbf{Correct}})$\end{tabular}} & \textbf{\begin{tabular}[c]{@{}c@{}}$\textbf{P}(\mathrm{\textbf{Chosen}}|\mathrm{\textbf{Incorrect}})$\end{tabular}} & \textbf{Lang} & \textbf{\begin{tabular}[c]{@{}c@{}}$\textbf{P}(\mathrm{\textbf{Chosen}}|\mathrm{\textbf{Correct}})$\end{tabular}} & \textbf{\begin{tabular}[c]{@{}c@{}}$\textbf{P}(\mathrm{\textbf{Chosen}}|\mathrm{\textbf{Incorrect}})$\end{tabular}} \\ \midrule
\multirow{4}{*}{\begin{tabular}[c]{@{}c@{}}Qwen2.5-\\ 72B\end{tabular}} & en & 42.3 & 39.3 & ar & 6.1 & 8.1\\
 & es & 37.8 & 32.4 & de & 36.2 & 36.5\\
 & ja & 21.6 & 15.1 & ja & 22.0 & 17.0\\
 & th & 7.4 & 6.2 & zh & 42.9 & 32.2 \\ \midrule
\multirow{4}{*}{\begin{tabular}[c]{@{}c@{}}LLaMA3.1-\\ 70B\end{tabular}} & fr & 41.9 & 38.2 & de & 20.1 & 21.5\\
 & ko & 27.6 & 26.6 & en & 54.3 & 45.4 \\
 & sw & 10.8 & 15.0 & hu & 10.8 & 13.6\\
 & vi & 19.6 & 19.7 & ru & 17.4 & 17.4 \\ \midrule
\multirow{4}{*}{\begin{tabular}[c]{@{}c@{}}R1-Distill-\\ LLaMA-70B\end{tabular}} & ar & 17.2 & 14.2 & ar & 13.5 & 8.2 \\
 & es & 50.2 & 40.0 & es & 37.7 & 32.6 \\
 & ko & 28.7 & 23.2 & ru & 26.5 & 13.7\\
 & sr & 16.4 & 8.5 & vi & 40.1 & 26.5 \\ \bottomrule
\end{tabular}
}

\caption{With LLM-as-a-judge answer selection, the models care more for the language instead of the correctness of the answer, and only R1-Distill-LLaMA revealed steady preference of the correct answers in all the judged languages. Language chosen rates in Multilingual LLM-as-a-judge answer selection, where $P(\mathrm{Chosen}|\mathrm{Correct})$ refers to the chosen rate of this language when its answer is correct, and $P(\mathrm{Chosen}|\mathrm{Incorrect})$ when the answer is incorrect. The tested languages are from the best-performing combinations in the experiments in \S\ref{sec:LLM_as_a_judge}.}
\label{tab:llm-judge-chosen-rate}
\end{table*}

Similarly, for LLM-as-a-judge selection, the judge models tend to prefer answers in high-resource languages, even if the answer of that language is incorrect. Table \ref{tab:llm-judge-chosen-rate} shows chosen rate of languages when the answer in that language is correct or incorrect. The results show that, only R1-Distill-LLaMA-70B showed stable preference of the correct answers among all the tested languages. Instead, all these models tend to roughly maintain the ratios (high for high-resource and low for low-resource) between languages regardless of the answer correctness. These results suggest that the models care more for the language instead of the correctness in judging, explaining why the Multilingual settings only beat Repeat and Paraphrase for R1-Distill-LLaMA-70B with LLM-as-a-judge selection.

\section{Conclusion}

In this paper, we comprehensively explore the benefits of multilingualism in reasoning and highlight several intriguing phenomena. Our findings suggest that utilizing multiple languages can significantly enhance reasoning capabilities, with a high upper bound for this benefit. Notably, this advantage is resilient to variations in translation quality and language choice, yet it remains sensitive to the methods used for answer selection. We examine various commonly used answer selection techniques but find that they often fall short of fully harnessing the potential of multilingualism in reasoning tasks. This disparity between the theoretical upper bound and practical experimental outcomes presents both a challenge and a promising avenue for future research.

\section*{Limitations}

In this study, while providing valuable insights into the potential of multilingualism in reasoning, has several notable limitations. However, our focus is primarily on large models with over 70 billion parameters, which may not fully represent the capabilities or challenges faced by smaller models. This narrow scope could lead to an incomplete understanding of how multilingualism affects reasoning across various architectures and sizes. Additionally, although we observe several interesting phenomena, the absence of a universal and stable method for leveraging multilingualism in reasoning.

\clearpage
\newpage
\normalem

\clearpage
\newpage

\appendix

\begingroup
\renewcommand{\arraystretch}{1.3}
\begin{table*}[!ht]
    \centering 
    \scriptsize
    \begin{tabular}{c|p{12cm}}
    \toprule
    \textbf{Setting} & \multicolumn{1}{|c}{\textbf{Prompt}} \\
    \midrule
    \rowcolor{lightgray!30}   \multicolumn{2}{c}{GPQA} \\
    \midrule
     \multirow{8}{*}{Default} & \textbf{System prompt:} Always think step by step and give your final choice among (A), (B), (C) and (D) by \"Answer: \{Your Choice\}\" in a single last line. \\

    & \makecell[l]{\textbf{User prompt:} What is the correct answer to this question:{\{Question\}}\\Choices:\\(A) {{choice1}}\\(B) {{choice2}}\\(C) {{choice3}}\\(D) {{choice4}}\\Let's think step by step:}  \\
    \midrule
    \multirow{8}{*}{\makecell{Prompt-Based Selection\\Translation = True}} & \textbf{System prompt:} Always choose the most suitable language, translate the question into that language, and think step by step in that language. Give your final choice among (A), (B), (C) and (D) by \"Answer: \{Your Choice\}\" in a single last line. \\
    & \makecell[l]{\textbf{User prompt:} What is the correct answer to this question:{\{Question\}}\\Choices:\\(A) {{choice1}}\\(B) {{choice2}}\\(C) {{choice3}}\\(D) {{choice4}}\\Let's think step by step:} \\
    \midrule
    \multirow{8}{*}{\makecell{Prompt-Based Selection\\Translation = False}} & \textbf{System prompt:} Always choose the most suitable language, and think step by step in that language. Give your final choice among (A), (B), (C) and (D) by \"Answer: \{Your Choice\}\" in a single last line. \\
    & \makecell[l]{\textbf{User prompt:} What is the correct answer to this question:{\{Question\}}\\Choices:\\(A) {{choice1}}\\(B) {{choice2}}\\(C) {{choice3}}\\(D) {{choice4}}\\Let's think step by step:} \\
    \midrule
    \midrule
    \rowcolor{lightgray!30}   \multicolumn{2}{c}{MGSM} \\
    \midrule
    \multirow{3}{*}{Default} & \textbf{System prompt:} Always think step by step and give your final answer by \"Answer: {Your Answer}\" in a single last line. \\
    & \makecell[l]{\textbf{User prompt:} Question: {\{Question\}}\\Step-by-Step Answer:} \\
    \midrule
    \multirow{3}{*}{\makecell{Prompt-Based Selection\\Translation = True}} & \textbf{System prompt:} Always choose the most suitable language, translate the question into that language, and think step by step in that language. Give your final answer by \"Answer: {Your Answer}\" in a single last line. \\
    & \makecell[l]{\textbf{User prompt:} Question: {\{Question\}}\\Step-by-Step Answer:}\\

    \midrule
    \multirow{3}{*}{\makecell{Prompt-Based Selection\\Translation = False}} & \textbf{System prompt:} Always choose the most suitable language, and think step by step in that language. Give your final answer by \"Answer: {Your Answer}\" in a single last line. \\

   & \makecell[l]{\textbf{User prompt:} Question: {\{Question\}}\\Step-by-Step Answer:} \\

    \bottomrule
    \end{tabular}
    \caption{The prompt template we used in experiments for each task.}
    \label{tab:prompt_template}
\end{table*}
\endgroup

\section{Models}
\label{sec:appendix_models}

\subsection{Model Description}
\paragraph{Qwen2.5-72B} is a cutting-edge language model designed to enhance natural language processing tasks with its impressive 72 billion parameters. This model excels in generating coherent and contextually relevant text, making it particularly valuable for applications in content creation, conversational agents, and automated summarization. 

\paragraph{LLaMA3.1-70B} represents the latest iteration in the LLaMA series, boasting 70 billion parameters that empower it to tackle complex reasoning tasks and generate high-quality text. This model is particularly noted for its ability to engage in multi-turn conversations, maintaining context and coherence over extended interactions.

\paragraph{R1-Distill-LLaMA-70B} is a distilled version of the original LLaMA model, optimized for efficiency without compromising performance. With 70 billion parameters, this model is designed to deliver faster response times and reduced computational requirements, making it ideal for deployment in resource-constrained environments.

\subsection{Languages-Related Prompt}

We present the prompt templates utilized in our experiments, including the Default and Prompt-based selection, as shown in Table~\ref{tab:prompt_template}. In the prompt-based selection experiments, we incorporated language-related constraints regarding whether to translate the question. Consequently, there are two variations of prompt-based selection: Translation=True and Translation=False, as indicated in the table.

\section{Results on MGSM}
\label{sec:appendix_mgsm}

We demonstrate the results of the three models on the MGSM task.
The results of the Repeat, Paraphrase, Multilingual, Repeat-Mix, and Paraphrase-Mix methods are presented in Table~\ref{tab:mgsm_human_repeat}, Table~\ref{tab:mgsm_human_paraphrase}, Table~\ref{tab:mgsm_human_multilingual}, Table~\ref{tab:mgsm_human_repeat_mix}, and Table~\ref{tab:mgsm_human_repeat_paraphrase}, respectively.
Table~\ref{tab:mgsm_google_multilingual} shows the results on the Google translated MGSM task.

\section{Used Scientific Artifacts}
Below lists scientific artifacts that are used in our work. For the sake of ethic, our use of these artifacts is consistent with their intended use.
\begin{itemize} [itemsep=1pt]
    \item \textit{LLaMA-3.1 (LLaMA3.1 license)}, a large language model developed by Meta. 
    \item \textit{R1-Distill-LLaMA-70B (MIT license)}, a large language model developed by Deepseek.
    \item \textit{Qwen-2.5-72B (Qwen license)}, a large language model developed by Qwen.
\end{itemize}

\begin{table*}[ht]
    \centering
    \footnotesize
    \begin{tabular}{c|c|cccc|ccc}
    \toprule
    \multirow{2}{*}{\textbf{Model}} & \multirow{2}{*}{\textbf{Setting}} & English & English & English & English & \multirow{2}{*}{$\overline{\textbf{Acc}}$} & \multirow{2}{*}{\textbf{Acc@4}} & \multirow{2}{*}{\textbf{Major@4}} \\
    & & Seed1 & Seed2 & Seed3 & Seed4 & & & \\
    \midrule
         \multirow{3}{*}{Qwen2.5-72B} & Best & 93.6 & 91.6 & 92.0 & 92.8 & 92.5 & 94.0 & 93.6 \\
         & Worst & 91.6 & 92.0 & 91.2 & 92.0 & 91.7 & 92.4 & 92.0 \\
         & Random & - & - & - & - & 92.4 & 93.7 & 92.5 \\
    \midrule
         \multirow{3}{*}{LLaMA3.1-70B} & Best & 91.6 & 92.0 & 92.4 & 93.6 & 92.4 & 96.0 & 93.2 \\
         & Worst & 90.0 & 90.8 & 91.6 & 91.6 & 91.0 & 92.4 & 91.2 \\
         & Random & - & - & - & - & 91.7 & 94.8 & 92.0 \\
    \midrule
        \multirow{3}{*}{R1-Distill-LLaMA-70B} & Best & 93.6 & 92.8 & 94.0 & 94.4 & 93.7 & 97.2 & 94.4 \\
        & Worst & 93.6 & 94.0 & 91.6 & 92.0 & 92.8 & 94.0 & 94.0\\
        & Random & - & - & - & - & 93.6 & 96.0 & 94.0 \\
    \bottomrule
    \end{tabular}
    \caption{The results of the Repeat method on the MGSM task.}
    \label{tab:mgsm_human_repeat}
\end{table*}

\begin{table*}[ht]
    \centering
    \footnotesize
    \setlength{\tabcolsep}{4pt}{
    \begin{tabular}{c|c|cccc|ccc}
    \toprule
    \multirow{2}{*}{\textbf{Model}} & \multirow{2}{*}{\textbf{Setting}} & English & English & English & English & \multirow{2}{*}{$\overline{\textbf{Acc}}$} & \multirow{2}{*}{\textbf{Acc@4}} & \multirow{2}{*}{\textbf{Major@4}} \\
    & & Paraphrase1 & Paraphrase2 & Paraphrase3 & Paraphrase4 & & & \\
    \midrule
         \multirow{3}{*}{Qwen2.5-72B} & Best & 92.0 & 88.8 & 90.8 & 91.6 & 90.8 & 96.0 & 93.2 \\
         & Worst & 89.6 & 88.4 & 88.0 & 88.4 & 88.6 & 92.4 & 90.0 \\
         & Random & - & - & - & - & 89.9 & 94.6 & 91.0 \\
    \midrule
         \multirow{3}{*}{LLaMA3.1-70B} & Best & 90.0 & 90.0 & 91.2 & 88.8 & 90.0 & 96.4 & 91.6 \\
         & Worst & 86.0 & 88.4 & 87.2 & 88.8 & 87.6 & 92.4 & 89.2 \\
         & Random & - & - & - & - & 88.9 & 94.8 & 90.8 \\
    \midrule
        \multirow{3}{*}{R1-Distill-LLaMA-70B} & Best & 89.6 & 91.2 & 91.2 & 90.4 & 90.6 & 96.8 & 91.6 \\
        & Worst & 89.6 & 89.2 & 89.2 & 89.2 & 89.3 & 92.8 & 92.0 \\
        & Random & - & - & - & - & 90.0 & 94.9 & 91.4 \\
    \bottomrule
    \end{tabular}}
    \caption{The results of the Paraphrase method on the MGSM task.}
    \label{tab:mgsm_human_paraphrase}
\end{table*}

\begin{table*}[ht]
    \centering
    \footnotesize
    \begin{tabular}{c|c|cccc|ccc}
    \toprule
    \textbf{Model} & \textbf{Setting} & Lang-1 & Lang-2 & Lang-3 & Lang-4 & $\overline{\textbf{Acc}}$ & \textbf{Acc@4} & \textbf{Major@4} \\
    \midrule
         \multirow{3}{*}{Qwen2.5-72B} & Best~(ar,cs,en,ko) & 91.2 & 84.8 & 92.4 & 91.2 & 89.9 & 98.0 & 94.0 \\
         & Worst~(bn,sw,te,zh) & 83.2 & 63.2 & 62.8 & 86.4 & 73.9 & 92.8 & 86.8 \\
         & Random & - & - & - & - & 84.7 & 95.8 & 91.7 \\
    \midrule
         \multirow{3}{*}{LLaMA3.1-70B} & Best~(ar,ru,sr,vi) & 87.2 & 90.0 & 87.2 & 90.0 & 88.6 & 99.6 & 92.8 \\
         & Worst~(bn,cs,sw,zh) & 81.2 & 85.2 & 84.4 & 84.4 & 83.8 & 93.6 & 90. \\
         & Random & - & - & - & - & 86.5 & 96.9 & 92.0 \\
    \midrule
        \multirow{3}{*}{R1-Distill-LLaMA-70B} & Best~(ar,bn,de,sr) & 90.8 & 75.2 & 86.4 & 92.4 & 86.2 & 99.6 & 92.8 \\
        & Worst~(bn,de,te,vi) & 75.2 & 86.4 & 78.0 & 87.6 & 81.8 & 95.6 & 90.4 \\
        & Random & - & - & - & - & 86.4 & 97.8 & 92.6 \\
    \bottomrule
    \end{tabular}
    \caption{The results of the Multilingual method on the MGSM task.}
    \label{tab:mgsm_human_multilingual}
\end{table*}

\begin{table*}[ht]
    \centering
    \footnotesize
    \begin{tabular}{c|c|cccc|ccc}
    \toprule
    \textbf{Model} & \textbf{Setting} & Lang-1 & Lang-2 & Lang-3 & Lang-4 & $\overline{\textbf{Acc}}$ & \textbf{Acc@4} & \textbf{Major@4} \\
    \midrule
         \multirow{3}{*}{Qwen2.5-72B} & Best~(en,en,es,te) & 92.4 & 92.8 & 90.8 & 62.8 & 84.7 & 97.6 & 93.2 \\
         & Worst~(en,en,ar,bn) & 92.4 & 92.8 & 91.2 & 83.2 & 89.9 & 93.2 & 92.8 \\
         & Random & - & - & - & - & 85.8 & 95.4 & 92.8 \\
    \midrule
         \multirow{3}{*}{LLaMA3.1-70B} & Best~(en,en,es,th) & 91.6 & 91.6 & 92.0 & 87.2 & 90.3 & 99.2 & 92.0 \\
         & Worst~(en,en,cs,de) & 91.6 & 91.6 & 85.2 & 88.8 & 89.3 & 92.8 & 91.2 \\
         & Random & - & - & - & - & 87.2 & 96.9 & 92.5 \\
    \midrule
        \multirow{3}{*}{R1-Distill-LLaMA-70B} & Best~(en,en,es,vi) & 92.8 & 93.2 & 87.2 & 87.6 & 90.2 & 99.6 & 94.0 \\
        & Worst~(en,en,ar,bn) & 92.8 & 93.2 & 90.8 & 75.2 & 88.0 & 94.8 & 93.2 \\
        & Random & - & - & - & - & 87.5 & 97.6 & 94.1 \\
    \bottomrule
    \end{tabular}
    \caption{The results of the Repeat-Mix method on the MGSM task.}
    \label{tab:mgsm_human_repeat_mix}
\end{table*}

\begin{table*}[ht]
    \centering
    \footnotesize
    \begin{tabular}{c|c|cccc|ccc}
    \toprule
    \textbf{Model} & \textbf{Setting} & Lang-1 & Lang-2 & Lang-3 & Lang-4 & $\overline{\textbf{Acc}}$ & \textbf{Acc@4} & \textbf{Major@4} \\
    \midrule
         \multirow{3}{*}{Qwen2.5-72B} & Best~(en,en,es,te) & 92.0 & 88.8 & 90.8 & 62.8 & 83.6 & 98.4 & 91.6 \\
         & Worst~(en,en,bn,cs) & 88.8 & 89.6 & 83.2 & 84.8 & 86.6 & 92.0 & 91.2 \\
         & Random & - & - & - & - & 85.4 & 95.7 & 91.8 \\
    \midrule
         \multirow{3}{*}{LLaMA3.1-70B} & Best~(en,en,es,te) & 90.0 & 90.0 & 92.0 & 82.8 & 88.7 & 99.2 & 93.6 \\
         & Worst~(en,en,ar,cs) & 90.0 & 90.0 & 87.2 & 85.2 & 88.1 & 92.8 & 91.6 \\
         & Random & - & - & - & - & 86.9 & 96.8 & 92.7 \\
    \midrule
        \multirow{3}{*}{R1-Distill-LLaMA-70B} & Best~(en,en,es,vi) & 89.6 & 91.6 & 87.2 & 87.6 & 89.0 & 99.6 & 93.2 \\
        & Worst~(en,en,cs,de) & 89.6 & 89.2 & 87.2 & 86.4 & 88.1 & 92.0 & 92.4 \\
        & Random & - & - & - & - & 86.8 & 96.7 & 92.7 \\
    \bottomrule
    \end{tabular}
    \caption{The results of the Paraphrase-Mix method on the MGSM task.}
    \label{tab:mgsm_human_repeat_paraphrase}
\end{table*}

\begin{table*}[ht]
    \centering
    \footnotesize
    \begin{tabular}{c|c|cccc|ccc}
    \toprule
    \textbf{Model} & \textbf{Setting} & Lang-1 & Lang-2 & Lang-3 & Lang-4 & $\overline{\textbf{Acc}}$ & \textbf{Acc@4} & \textbf{Major@4} \\
    \midrule
         \multirow{4}{*}{Qwen2.5-72B} & Best~(ar,en,es,hu) & 88.8 & 92.4 & 89.2 & 78.8 & 87.3 & 98.0 & 92.4 \\
         & Worst~(bn,cs,fr,sw) & 29.2 & 82.4 & 82.4 & 62.8 & 64.2 & 92.4 & 84.0 \\
         & Human~(ar,cs,en,ko) & 88.8 & 82.4 & 92.4 & 88.0 & 87.9 & 96.4 & 92.4 \\
         & Random & - & - & - & - & 80.5 & 96.0 & 90.7 \\
    \midrule
         \multirow{4}{*}{LLaMA3.1-70B} & Best~(de,fr,ja,vi) & 88.0 & 82.8 & 83.6 & 89.2 & 85.9 & 99.2 & 90.4 \\
         & Worst~(bn,cs,sw,zh) & 81.6 & 84.0 & 79.6 & 84.8 & 82.5 & 94.0 & 90.0 \\
         & Human~(ar,ru,sr,vi) & 86.4 & 89.2 & 85.6 & 89.2 & 87.6 & 97.2 & 93.6 \\
         & Random & - & - & - & - & 85.0 & 96.8 & 91.4 \\
    \midrule
        \multirow{4}{*}{R1-Distill-LLaMA-70B} & Best~(ar,hu,sr,zh) & 89.6 & 82.0 & 87.2 & 88.4 & 86.8 & 98.8 & 94.0 \\
        & Worst~(de,sw,te,vi) & 86.8 & 80.0 & 77.2 & 84.8 & 82.2 & 93.6 & 91.2 \\
        & Human~(ar,bn,de,sr) & 89.6 & 52.8 & 86.8 & 87.2 & 79.1 & 98.0 & 92.8 \\
        & Random & - & - & - & - & 83.6 & 97.2 & 92.1 \\
    \bottomrule
    \end{tabular}
    \caption{The results of the Multilingual method on the Google translated MGSM task.}
    \label{tab:mgsm_google_multilingual}
\end{table*}

\begin{table*}[ht]
\centering
\footnotesize
\begin{tabular}{c|cccc}
\hline
\textbf{Model} & \textbf{Langs} & \textbf{Acc@k} & \textbf{Major@k} & \textbf{Judge@k} \\ \hline
 & repeat & 93.6 & 93.2 & 91.2 \\
 & paraphrase & 94.0 & 91.6 & 90.0 \\
\multirow{-3}{*}{\begin{tabular}[c]{@{}c@{}}Qwen2.5-\\ 72B\end{tabular}} & ar,en,es,hu & 98.0 & 92.4 & 89.6 \\ \hline
 & repeat & 94.8 & 92.0 &  92.0 \\
 & paraphrase & 95.2 & 91.6 & 91.6 \\
\multirow{-3}{*}{\begin{tabular}[c]{@{}c@{}}LLaMA3.1-\\ 70B\end{tabular}} & de,fr,ja,vi & 99.2 & 90.4 & 88.0 \\ \hline
 & repeat & 96.4 & 93.6 & 92.8 \\
 & paraphrase & 93.6 & 91.6 & 88.8 \\
\multirow{-3}{*}{\begin{tabular}[c]{@{}c@{}}R1-Distill-\\ LLaMA-70B\end{tabular}} & ar,hu,sr,zh & 98.8 & 94.0 & 91.6 \\ \hline
\end{tabular}
\caption{LLM-as-a-judge performance on MGSM dataset}
\label{tab:llm-judge-math}
\end{table*}

\end{document}